\documentclass[journal]{vgtc}                     

\onlineid{0}

\vgtccategory{Research}

\usepackage{flushend}

\newcommand{\our}{LLMMap\xspace}
\newcommand{\ours}{LLMMaps\xspace}
\newcommand{\qna}{Q\&A\xspace}

\vgtcpapertype{please specify}

\title{LLMMaps - A Visual Metaphor for\\ Stratified Evaluation of Large Language Models}

\author{%
  Patrik Puchert,
  Poonam Poonam,
  \authororcid{Christian van Onzenoodt}{0000-0002-5951-6795},
  \authororcid{Timo Ropinski}{0000-0002-7857-5512}
}

\authorfooter{
  \item
  	Patrik Puchert is with Ulm University.
  	E-mail: patrik.puchert@uni-ulm.de
  \item
  	Poonam Poonam is with Ulm University.
  	E-mail: poonam.poonam@uni-ulm.de
  \item
  	Christian van Onzenoodt is with Ulm University.
  	E-mail: christian.van-onzenoodt@uni-ulm.de
  \item
  	Timo Ropinski is with Ulm University.
  	E-mail: timo.ropinski@uni-ulm.de
}

\abstract{Large Language Models (LLMs) have revolutionized natural language processing and demonstrated impressive capabilities in various tasks. Unfortunately, they are prone to hallucinations, where the model exposes incorrect or false information in its responses, which renders diligent evaluation approaches mandatory. While LLM performance in specific knowledge fields is often evaluated based on question and answer (Q\&A) datasets, such evaluations usually report only a single accuracy number for the dataset, which often covers an entire field. This field-based evaluation, is problematic with respect to transparency and model improvement. A stratified evaluation could instead reveal subfields, where hallucinations are more likely to occur and thus help to better assess LLMs' risks and guide their further development. To support such stratified evaluations, we propose LLMMaps as a novel visualization technique that enables users to evaluate LLMs' performance with respect to Q\&A datasets. LLMMaps provide detailed insights into LLMs' knowledge capabilities in different subfields, by transforming Q\&A datasets as well as LLM responses into an internal knowledge structure. An extension for comparative visualization furthermore, allows for the detailed comparison of multiple LLMs. To assess LLMMaps we use them to conduct a comparative analysis of several state-of-the-art LLMs, such as BLOOM, GPT-2, GPT-3, ChatGPT and LLaMa-13B, as well as two qualitative user evaluations. All necessary source code and data for generating LLMMaps to be used in scientific publications and elsewhere is available on GitHub: \url{https://github.com/viscom-ulm/LLMMaps}}


\teaser{
    \centering
    \label{fig:dummy}
    \includegraphics[trim=10 20 10 20, clip, width=\textwidth]{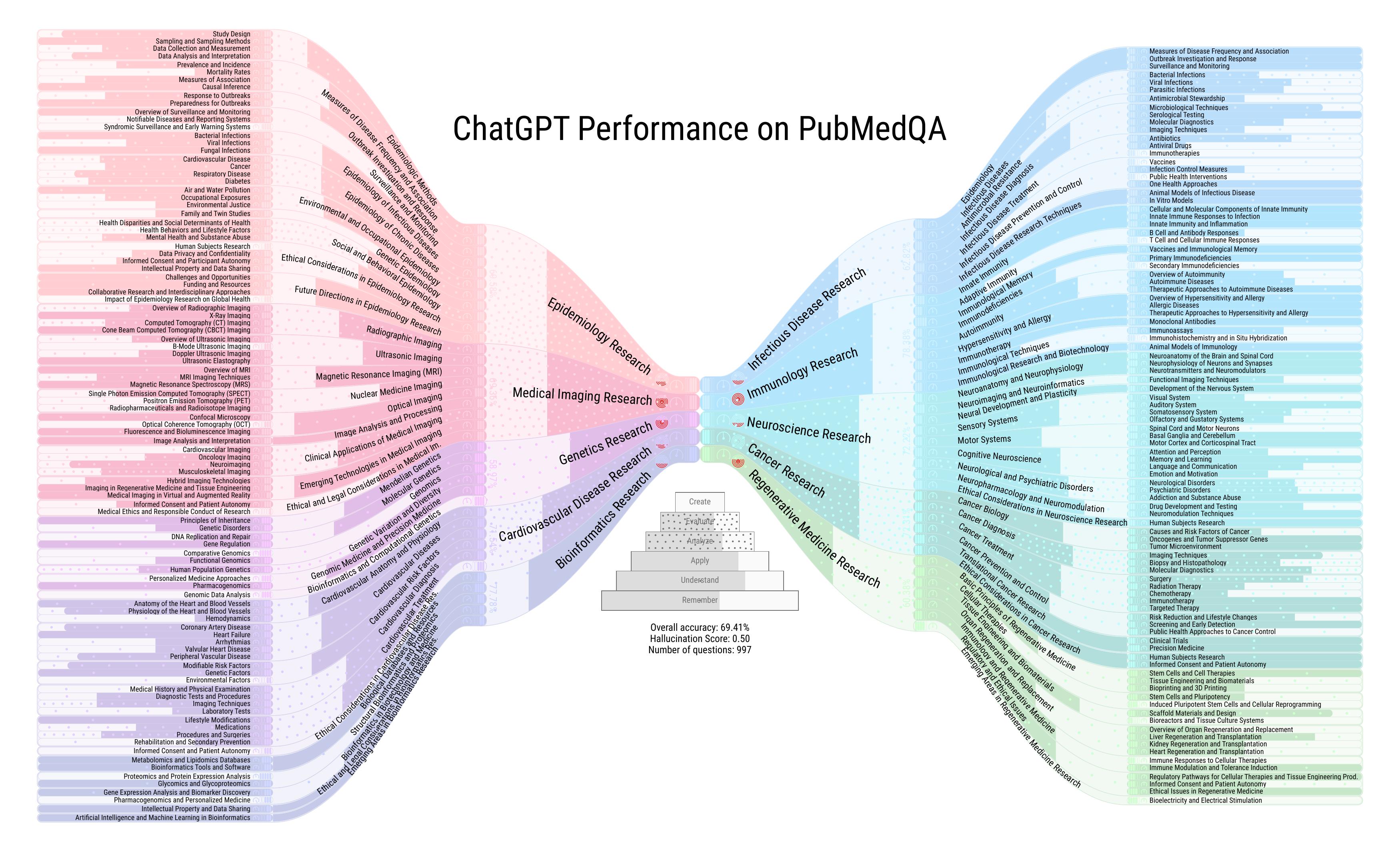}
    \caption{\our for ChatGPT's knowledge capability evaluation on PubMedQA, a \qna dataset containing 997 biomedical research questions. By exploiting knowledge stratification, we can visualize performance per subfields, whereby bars depict accuracy and blue noise dots depict the number of questions per subfield. Below, the performance is also stratified according to Bloom's learning dimension hierarchy. The canvas-like tree visualization is inspired by icicle plots, and our implementation allows for the integration of various details through a plugin mechanism. To demonstrate this feature, we have integrated estimations for difficulty levels, response times, and hallucination probability.}
    \label{fig:teaser}
}

\graphicspath{{figs/}{figures/}{pictures/}{images/}{./}} 

\usepackage{tabu}                      
\usepackage{booktabs}                  
\usepackage{lipsum}                    
\usepackage{mwe}                       

\usepackage{mathptmx}                  

\begin{document}


\firstsection{Introduction}

\maketitle

The advent of Large Language Models (LLMs) has been a significant breakthrough in the field of natural language processing (NLP) and has shown impressive capabilities in various tasks, such as language modeling, sentiment analysis, question-answering, and machine translation. LLMs are trained on massive text datasets that contain vast amounts of language data, which makes them perform well on tasks that require the understanding of the context and the ability to generate coherent and meaningful responses. Recently, researchers were able to show that state-of-the-art LLMs pass law exams~\cite{choi2023chatgpt} or even the United States Medical Licensing Examination (USMLE)~\cite{nori2023capabilities}. However, despite their impressive performance, LLMs also face several challenges and imply dangers~\cite{bender2021dangers}, including the problem of hallucination~\cite{ji23hallucinations}, which is a common phenomenon where the model includes incorrect or false information in its responses~\cite{weidinger2021ethical,weidinger2022taxonomy}, while still providing an eloquent answer~\cite{mahowald2023dissociating}. Unfortunately, this behavior is especially problematic in the most promising application domains, such as healthcare or legal settings, where incorrect information can have serious consequences. Due to these challenges, the evaluation of LLMs is mandatory, and today many benchmarks and leader boards exist.

Evaluating the performance of LLMs on large question and answer (\qna) datasets is a common practice in NLP and machine learning. In this evaluation method, the LLM is presented with a set of questions and is expected to generate correct responses, which are known via annotations but not revealed to the LLM via labels. While initially these question-based evaluations were tailored to investigate an LLM's language capabilities~\cite{devlin2018bert}, recently we can see a clear trend towards the investigation of knowledge capabilities~\cite{brown2020language,choi2023chatgpt,kung2023performance,openai2023gpt4,nori2023capabilities,bubeck2023sparks}. Based on the answered questions, the performance of the LLM is measured by computing the accuracy of its responses. While this method provides a quick and easy way to evaluate the accuracy of LLMs, it has limitations and does not provide a comprehensive understanding of the model's capabilities and limitations. Rather a more stratified analysis of LLMs is necessary to identify knowledge fields where it performs good or bad, and thus assess their performance in different subfields. Such a stratified evaluation not only contributes to the transparency of LLMs, but also enables developers to further improve their performance. For instance, by identifying subfields where LLMs underperform, LLM developers could collect additional training data or steer human reinforcement learners towards these subfields in order to inspect LLM answers more critically or simply more often. Moreover, a stratified evaluation could also be used to compare the performance of several LLMs and thus allow for a comparative evaluation on a subfield level, or even to gain insights to orchestrate ensemble LLMs based on a specific prompt.

To support a stratified analysis of LLMs, we propose \ours, an extendable visualization metaphor that enables LLM users and developers alike to evaluate the performance of one or multiple LLMs with respect to one or multiple \qna datasets (see Figure~\ref{fig:teaser}). Thus, \ours rather than showing an aggregated accuracy, enable novel and detailed insights regarding the knowledge capabilities of LLMs in different knowledge subfields. Towards this end we make the following contributions within this paper:

\begin{itemize}
\item We present \ours as a carefully designed, extendable visualization technique to support a stratified and comparative evaluation of LLM knowledge capabilities on \qna datasets.
\item We suggest how to inform \ours by knowledge stratification strategies for \qna datasets, in order to derive an underlying knowledge hierarchy.
\item We conduct several qualitative evaluations highlighting the benefits and downsides of \ours.
\item We have released all source code and data, to enable both, NLP researchers to use, and VIS researchers to further improve \ours.
\end{itemize}

\section{Related Work}\label{sec:relatedwork}

\noindent\textbf{LLM evaluation.} Evaluating the performance of LLMs is a crucial step in developing and improving their capabilities. To consider the state of the art on LLM evaluation, we have reviewed recent publications on LLMs, whereby we had a special focus on the evaluation of the models' knowledge capabilities. While these publications consist of several works, which propose and evaluate a single LLM, there have also been frameworks proposed for the comparative evaluation of multiple LLMs.

One common evaluation practice in NLP is to test LLMs on large \qna datasets. In this method, the model is presented with a set of questions and is expected to generate correct responses without access to the annotated labels. Typically, obtained LLM responses are evaluated using metrics that aim to measure their performance in language-related tasks~\cite{liang2022holistic,lin-2004-rouge}. However, for knowledge assessment, accuracy and F1-score are more relevant. To quantify results on \qna datasets, accuracy is often simply considered as exact-match accuracy, i.e., the model generated response must exactly match one of the optional answers. This can be further expanded to a quasi-exact match score to focus on the knowledge contained in the answer rather than the exact formulation. For other tasks with an even broader output space, the accuracy becomes more difficult to quantify. For question answering tasks, the F1-score is commonly used to measure the overlap of words in desired and given responses. Furthermore, MRR and NDCG scores~\cite{10.1145/582415.582418} can be applied to measure information retrieval in the LLMs generated output. For all these metrics, the accuracy of the model's responses is then calculated as the ratio of the model's correct answers to the total number of questions. While these metrics provide a quick and easy way to evaluate the accuracy of LLMs, they do not help to obtain a comprehensive understanding of their capabilities and limitations. 

 Were language models traditionally evaluated based on language understanding tasks, the observation that LLMs are few-shot learners, which arose with the advent of GPT-3~\cite{brown2020language}, lead to this shift in the used evaluation benchmark datasets. While the original BERT model~\cite{devlin2018bert} was evaluated on language understanding benchmark datasets, such as the General Language Understanding Evaluation (GLUE) benchmark~\cite{wang2018glue}, the Stanford Question Answering Datasets (SQuAD)~\cite{rajpurkar2016squad}, and the Situations With Adversarial Generations benchmark (SWAG)~\cite{zellers2018swagaf}, a shift to knowledge based \qna datasets is observable in modern LLMs. GPT-3~\cite{brown2020language} is for instance evaluated on the TriviaQA dataset~\cite{JoshiTriviaQA2017}, law exams~\cite{choi2023chatgpt}, and medical questionnaires~\cite{kung2023performance}, while GPT-4 has been evaluated on more than 30 academic and professional exams~\cite{openai2023gpt4} as well as medical questionnaires~\cite{nori2023capabilities}. Bubeck et al. go even a step further, and not only evaluate GPT-4's knowledge capabilities, but also its general intelligence~\cite{bubeck2023sparks}. This trend is also captured by modern evaluation frameworks, such as Stanford's Holistic Evaluation of Language Models (HELM)~\cite{liang2022holistic} or OpenAI's recently initiated Eval framework.
 
 LLM evaluation is also important for model deployment. Mitchel et al. call for new technology to better communicate risks of deployed language models~\cite{mitchell2019model}. Their proposed model cards, should be released together with a model, and contain details regarding the model's benchmarked evaluation characteristics. Similarly to as we see \ours, Mitchell et al. see model cards as a step towards the responsible democratization of ML models, by increasing transparency into how well models perform. Raj et al. go even a step further, and demand specific AI audit trails, which also contain model cards, but are meant as an internal audit framework to assess risks of AI models before their deployment~\cite{raji2020closing}. While some researchers explicitly underline the importance of visual representations in model cards~\cite{crisan2022interactive}, LLM evaluations rarely make use of visualizations beyond simple bar charts, e.g.,~\cite{brown2020language,touvron2023llama,scao2022bloom,taori23alpaca,openai2023gpt4,liu2303g}. Based on these observations, we see a clear need for visualization in the context of LLM evaluation, even beyond \ours.

\noindent\textbf{LLM visualization.} Burkhard defines knowledge visualization as the use of visual representations to facilitate knowledge transfer between two or more persons~\cite{burkhard2004learning}. Thus, it must not only visualize the available information, but boil down to what is important. While general approaches like knowledge graphs~\cite{hogan2021knowledge} exploit every relation in the data to generate a visual representation, this representation is overly extensive to be of help to non-experts. To this end, more stratified approaches can help to convey the desired knowledge in an easily understandable and efficient way. Popular visualizations of such stratified, and thus hierarchical, data can take the form of dendrograms, radial tree maps or mindmaps~\cite{schulz2011treevis}. In the context of LLMs this has been primarily applied to the underlying data, by visualizing sentence structures~\cite{wang2022self}. Even though this can be of great help for developers and prompt designers, it does not help in conveying knowledge capabilities.

Other applications of visualization on LLMs focus on aspects of explainable AI to make the decision making process of AI models understandable for humans. Usually, these approaches are addressed towards experts, as the understandability of what leads the network to a certain decision, requires also an understanding of its functioning. Some works focus on how the network actually achieves a certain result by exploring the textual and linguistic structure of the generated output~\cite{li2015visualizing}, by visualizing the inner workings of the network utilizing saliency and attention maps~\cite{wang2021dodrio}, or by highlighting the contextualization of word embeddings~\cite{sevastjanova2022lmfingerprints,sevastjanova2022visual}.

Although, these approaches are of great help to developers, they do not leverage the understandability of an LLM's accuracy in the targeted knowledge fields.

\section{Design Considerations}\label{sec:designconsiderations}
Within this section, we briefly discuss the considerations, which underlie or visualization design introduced in Section.~\ref{sec:visdesign}. The more technically inclined readers mainly interested in the visualization aspects of \ours might skip this section for now, and refer to it later when diving deeper into our work.

Since the knowledge capabilities of LLMs are progressing fast, and to some extent towards those of a human, we take design inspiration from the current state of the art in knowledge assessment as done in education. Knowledge assessment is a process for determining the nature and extent of human learning and development~\cite{nitko1996educational}. As modern LLMs, such as GPT-4 are quoted as exhibiting human-level performance on various professional and academic benchmarks~\cite{openai2023gpt4}, grounded knowledge assessment becomes more and more important in the context of LLMs. When addressing knowledge assessment on a high level, we can differentiate between formative and summative assessment. As summative assessment in education is concerned with learning outcomes, we also apply this paradigm for LLM evaluation. It is widely considered to be effective and useful when knowledge assessment is integrated across multiple settings such as targeting it towards individual courses of a curriculum~\cite{suskie2018assessing}. For the individual courses again, it is often helpful to assess knowledge for different subfields, such as for instance the chapters of a lecture course. Humans might be doing well in some subfields, while performing subpar in others. We see this as a further motivation for supporting stratification in the context of LLM evaluation.

For effective assessment in LLMs, we should further focus on important learning goals and take into account what is considered evidence for human learning. From the main groups of evidence for human learning, such as written work, performance, and presentation~\cite{nitko1996educational}, we see performance as measured based on correct test responses, as the most obvious direct evidence for LLM \emph{learning}. While written work could certainly also be considered, it is more difficult to rate, as it mixes text generation and knowledge capabilities~\cite{mahowald2023dissociating}. Therefore, we and many other researchers~\cite{brown2020language,kung2023performance,openai2023gpt4,nori2023capabilities,bubeck2023sparks}, focus on LLM knowledge assessment based on \qna datasets.

Also, as the existing LLM evaluations inspiring our visualization design can mainly be found in scientific publications or on web pages, we have designed \ours to be usable in these contexts, and ideally beyond.

\section{Visualization Design}\label{sec:visdesign}
In this section, we will detail our visualization design, carefully crafted according to the design considerations discussed in Section~\ref{sec:designconsiderations}.

\subsection{Visualization Layout}
Our main goal is to provide a visual and extendable canvas, which serves the evaluation of LLMs. To support stratified knowledge-based visualization, our main visualization layout employs the structural benefits of mindmaps, and takes inspiration from classical icicle plots~\cite{kruskal1983icicle}. Mindmaps are not only used for problem-based learning, small-group teaching, as well as in examination tools~\cite{edwards2010mind}, but it could also be shown that they engage learners~\cite{chiou2008effect,selvi2018case} and positively affect learning outcomes in different domains~\cite{brinkmann2003graphical,d2010does}. With their explicit representation of the knowledge hierarchy, we also assume that the interpretation of \ours does not require lengthy explanations. We further exploit an axis-parallel alignment of the mindmap, as opposed to a radial alignment, as we consider it a better fit for the usage of \ours in scientific publications.

\begin{figure*}[h]
\centering
    \begin{tabular}{ccc}
        \includegraphics[scale=0.8]{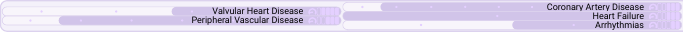}&
        \includegraphics[scale=0.8]{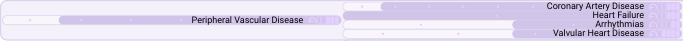}&
        \includegraphics[scale=0.8]{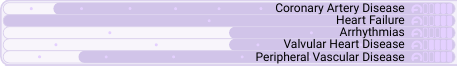}\\
        \includegraphics[width=0.3\textwidth]{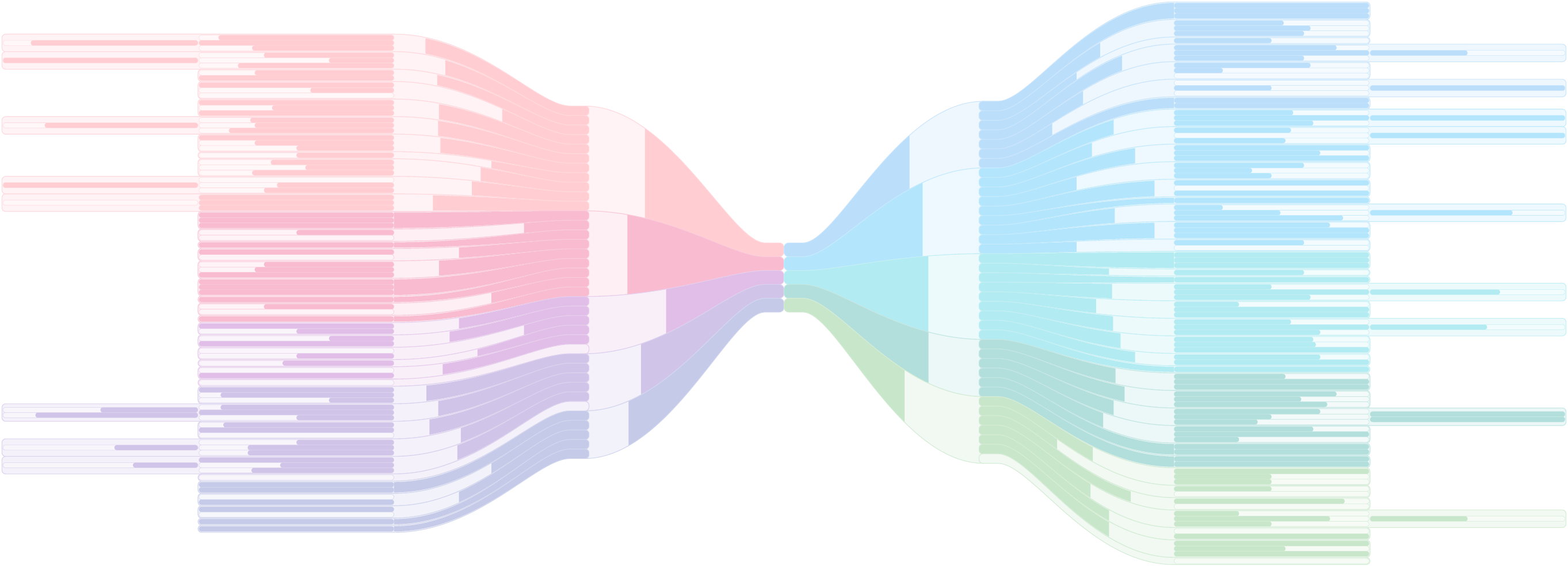}&
        \includegraphics[width=0.3\textwidth]{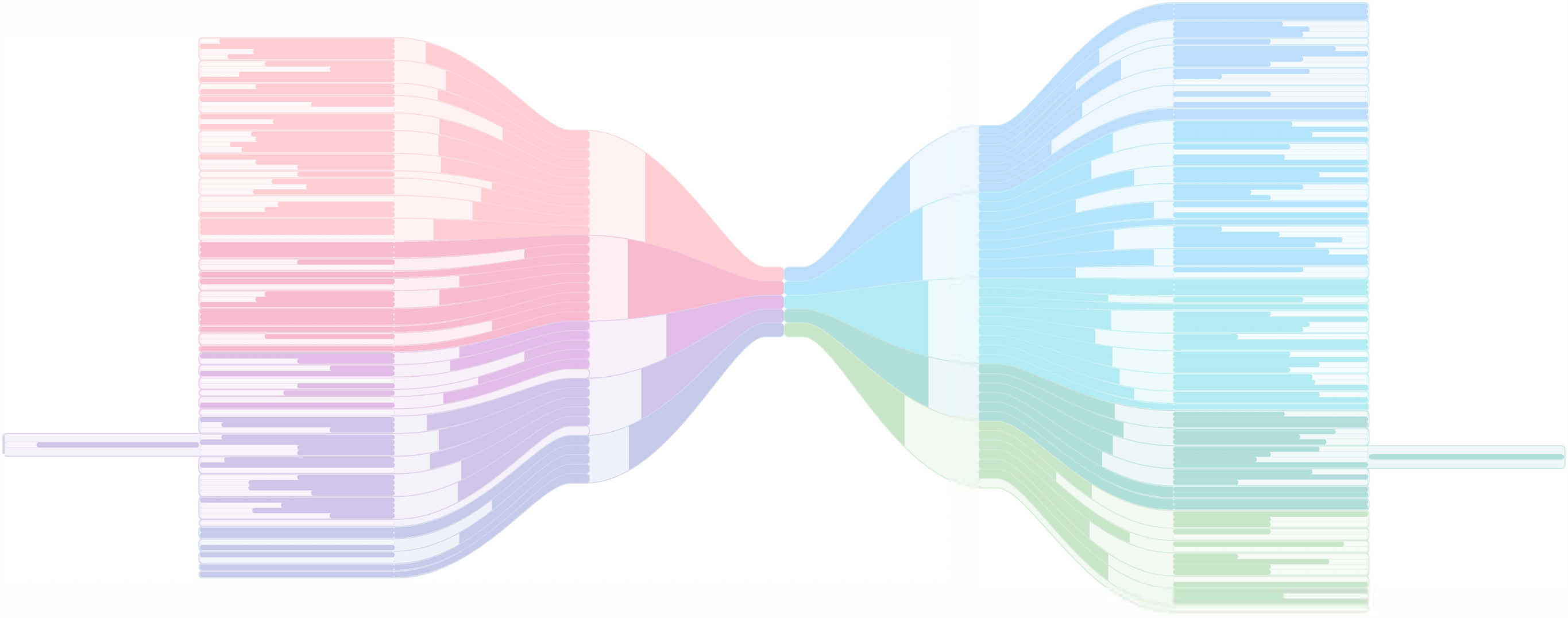}&
        \includegraphics[width=0.225\textwidth]{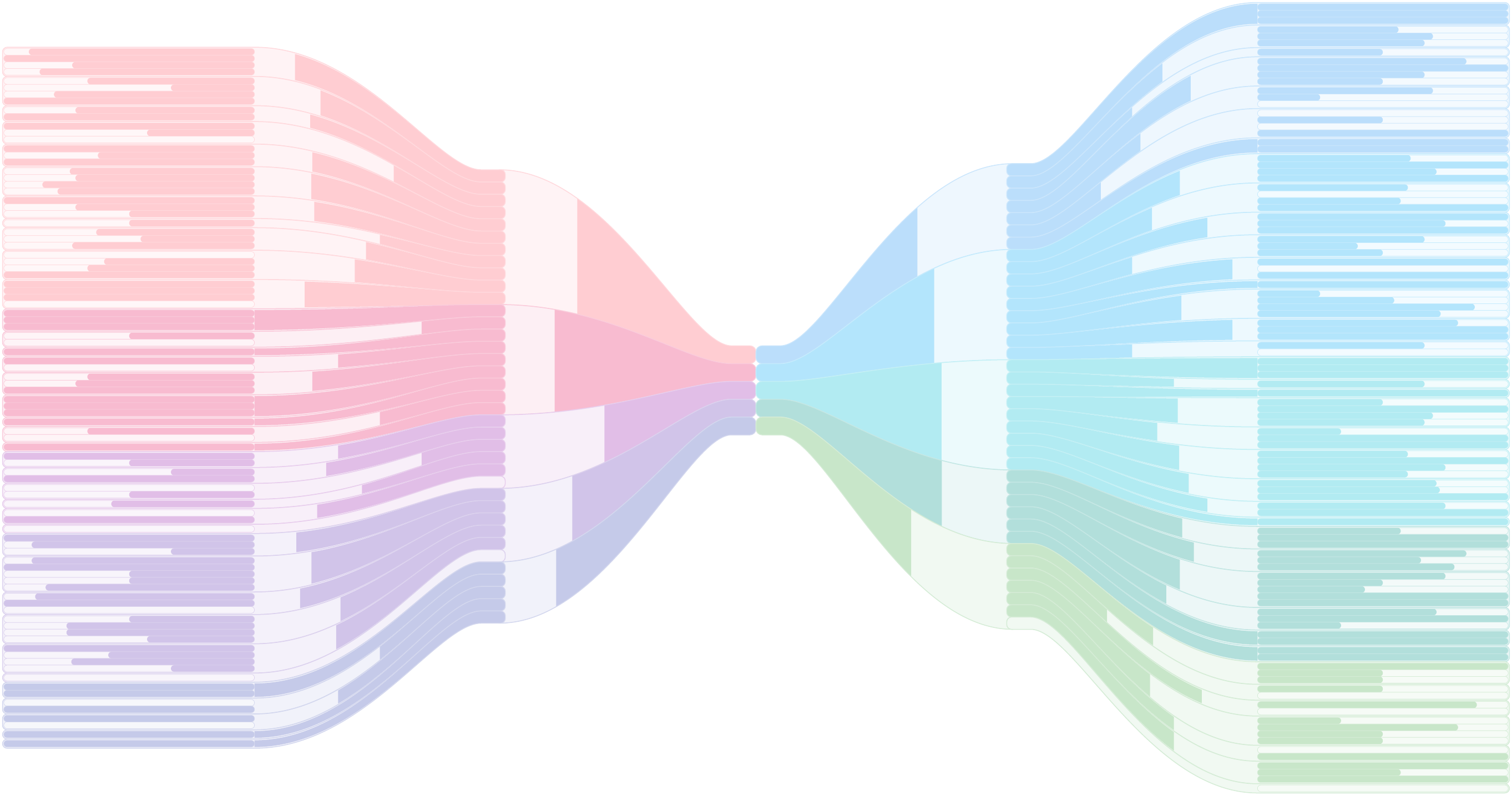}\\
        (a) $h_{ln}=3$ & (b) $h_{ln}=4$ & (c) $h_{ln}=5$\\
    \end{tabular}
    \caption{Comparison of different values for leaf stacking parameter $h_{ln}$. Larger values for $h_{ln}$ result in more leaf nodes stacked (\emph{top row}), which trades overall figure width for height (\emph{bottom row}). Parent-child relations on leaf level are emphasized by parents encapsulating child nodes.}
    \label{fig:leafstacking}
\end{figure*}

The overall layout of our visualization canvas is informed by a knowledge hierarchy, which is derived from \qna datasets, and which is given by a tree data structure. The tree data structure hierarchically structures a knowledge field, given by a \qna data set, and assigns the containing questions to the leaf nodes. Section~\ref{sec:knowledgestratification} suggests approaches for generating this underlying data structure. When visualizing such a knowledge hierarchy in a horizontal axis-parallel layout, we have experienced figure height, as opposed to figure width, as the main constraining factor for all handled \qna datasets. Therefore, we have decided to split the knowledge tree at the root node into halves and visualize one half oriented towards the left and one towards the right side. This design decision is in line with Ondov et al.'s findings, which report cases with significant task performance improvements when arranging small multiples in a mirror-symmetric fashion~\cite{ondov2018face}. While this layout reduces figure height roughly by 50\%, figure height still remained the major limiting factor. Therefore, we have further decided to be able to horizontally stack leaf nodes in such a way, that we are able to exploit the full page width, while at the same time reducing the figure height. To realize this leaf node stacking, we introduce a user parameter $h_{ln}$, which specifies the maximum number of leafs that are visualized on top of each other for a given parent. Figure~\ref{fig:leafstacking} illustrates this concept for $h_{ln}={3,4,5}$. As can be seen, the stacked leaf nodes are encapsuled by their parent nodes, to better communicate parent-child relations. Since we found that $h_{ln}=3$ worked well for all handled \qna datasets we decided to set this as the default value.

Another important consideration, which affects figure size, is whether to show empty leaf nodes, i.e., those subfields for which the \qna dataset does not contain any questions. For the knowledge stratification strategies suggested in Section~\ref{sec:knowledgestratification}, we found that the handled \qna datasets resulted in the following percentages of empty leaf nodes: PubMedQA~\cite{jin2019pubmedqa} 51.00\%, MedMCQA~\cite{pmlr-v174-pal22a} 30.08\%, SciQ~\cite{welbl-etal-2017-crowdsourcing} 21.13\%, and USBar~\cite{ncbe_usbar} 19.12\%. Especially for PubMedQA and MedMCQA the rather high number of empty leaf nodes, has a significant contribution to overall figure height, especially for low $h_{ln}$. Empty leaf nodes play an important role, when analyzing \qna datasets, as they indicate subfields which are not represented by the dataset. However, when the main goal is to see how a LLM performs on a given benchmark dataset, we consider them less relevant. Therefore, we have decided to omit empty leaf nodes per default from our visualization, while still giving users the opportunity to override this with a dedicated user parameter.

\begin{figure}[b]
    \centering
    \begin{subfigure}{0.45\linewidth}
        \centering
        \includegraphics[trim=500 280 350 280, clip, width=0.95\linewidth]{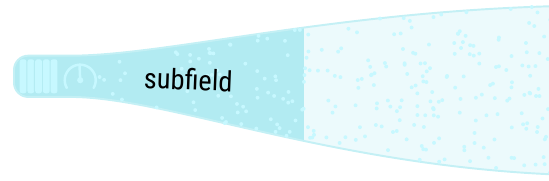}
        \caption{Number of questions per subfield represented as random noise pattern}
        \label{fig:bluenoise:a}
    \end{subfigure}
    \begin{subfigure}{0.45\linewidth}
        \centering
        \includegraphics[trim=500 280 350 280, clip, width=0.95\linewidth]{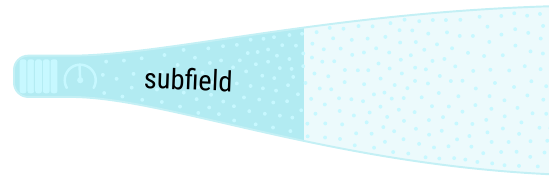}
        \caption{Number of questions per subfield represented as blue noise pattern}
        \label{fig:bluenoise:b}
    \end{subfigure}
    \caption{Comparison of randomly distributed points \emph{(a)} vs. blue noise distributed points \emph{(b)} for encoding the number of questions per subfield. Due to the blue noise criterion no point aggregations influence the perceived density.}
    \label{fig:bluenoise}
\end{figure}

\subsection{Quantity Encoding}
While the visualization layout described in the previous subsection generates the main visualization canvas, it does not provide any quantitative values regarding an LLM's performance. As the LLM performance per subfield is the most important information to be communicated by \ours, our intention was to bring across this information as prominent and as clearly as possible. Thus, we have decided to exploit bar charts for this purpose, as we visualize accuracy per subfield by means of a horizontal bar -- another inspiration taken from classical icicle plots~\cite{kruskal1983icicle}. While this enables direct performance comparison within each half of the knowledge tree, comparison across halves is less direct. Furthermore, a bar does not provide exact quantitative values without the presence of appropriate axis labels. We therefore have decided to also directly display the accuracy as number. Since, we see this only as a secondary means of performance encoding, we have traded contrast for overall visualization design and have used the primary subfield color for this text label. Thus we are able to generate \ours as for instance shown in Figure~\ref{fig:teaser}.

While Figure~\ref{fig:teaser} supports the stratified evaluation of a single LLM, it does not allow for performance comparisons of multiple LLM's based on one or several \qna datasets. To also support this use case scenario, we have extended \ours by being able to communicate the performance of multiple models. To do so, we horizontally split the performance bar into $k$ bars, where $k$ is the number of visualized LLMs. Furthermore, while for the single-model use case scenario, we have chosen hue, as the most discriminative color feature, to visually encode subfield, we have decided to use hue to visually encode model in our multi-model scenario. Figure~\ref{fig:multimodel} shows an example of the resulting visualization. As can be seen, it now becomes possible to compare models per subfield, while still being able to compare their overall performance through the conventional bar chart used in most LLM performance comparisons.

While the accuracies per subfield are important, their relevance for the overall model performance on a given \qna dataset is only interpretable, when considering the actual number of questions associated with a specific subfield. Therefore, we also have to visualize this quantity, whereby we consider it of subordinate importance as opposed to the performance score. As the number of questions per subfield can be interpreted as a density value, we decided to choose commonly used density visualizations to display it. As point-based density plots are both widely used, and non-conflicting with the used bar chart, we decided to visualize the number of questions as dot plot. Since we have both dimensions within a subfield as free variables, we have decided to realize our dot plots as an extension to Blue Noise Plots~\cite{van2021blue}. By facilitating a blue noise distribution, we are able to evenly distribute the points within each subfield, without resulting in unwanted clustering effects. To do so, we have extended the blue noise plot algorithm to support point distribution within arbitrary shapes, as given by our subfield regions. During point relaxation, our approach computes on which side of a closed path consisting of multiple cubic Bezier curves a point lies. Once a point is outside of this path, we move it back along the direction normal onto the path. Doing so ensures that we obtain equally distributed points within each subfield. Figure~\ref{fig:bluenoise} shows a comparison of our blue noise distribution as compared to a unconstrained random distribution, where overlap and clustering affects the perception of question density.

\subsection{Color Coding}
As in all visualization scenarios, an appropriate color coding is mandatory to obtain an interpretable visualization. Therefore, we have developed two alternative color coding strategies, which have been designed with the single- and multi-model evaluation in mind.

\noindent\textbf{Data-centric color coding.} Our data-centric color coding approach is meant for single-model evaluations. In this case, we have chosen a color scheme, which allows for differentiating the different subfields by employing Google's material colors. For each subfield, we use a darker shade for the accuracy bar, whereby the subfield is visualized with a brighter shade (see Figure~\ref{fig:teaser}).

\noindent\textbf{Model-centric color coding.} The model-centric color coding is meant to be used for multi-model evaluations, where it is important to clearly associate each bar with a specific model. Therefore, we use hue, as the most dominant color parameter, to depict models, rather than subfields (see Figure~\ref{fig:multimodel}). To also allow for differentiating the different subfields we use alternating shades for their background color.

\subsection{Meta Information}
To interpret \ours, the integration of global information by means of title and textual legend is also essential. So, for each \our we display a comprehensive title detailing which model(s) has/have been evaluated on which dataset(s), and we display the overall score of the visualized models.

\section{Knowledge Field Stratification}\label{sec:knowledgestratification}
In this section, we will suggest strategies to generate the knowledge hierarchies used by \ours. While this is not the main focus of our paper, such hierarchies are still important for enabling \ours, and we have designed the released source code, such that future users and developers can exploit their own hierarchies. 

Most \qna datasets are either not or only very coarsely stratified. Some contain 1 out of only 5-10 different topic labels associated with each question~\cite{hendrycks2020measuring,pmlr-v174-pal22a}. In other cases, such labels are so fine-grained~\cite{jin2019pubmedqa} that using them to define subfields would clutter the visualization. Sometimes no subfield related classification labels are available, e.g., a measured difficulty for the publicly available US Multistate Bar Examination questions, which do however not stratify the knowledge fields. While all question classification labels we found in the processed \qna datasets make knowledge stratification difficult, the observed dataset differences further complicate matters, and we see a need for a robust and generalizable knowledge hierarchy derivation for \qna datasets.

To address this challenge, we propose a top-down knowledge stratification strategy, which we developed to associate individual questions with the leaf nodes of the obtained knowledge hierarchy. Our approach starts by identifying the overarching subfields of the knowledge field represented by a \qna dataset. Afterwards each of these overarching subfields, is further divide into finer-grained subfields, which we recursively divide again until the target depth of the resulting knowledge tree is reached. Finally, each question is associated with the leaf node best reflecting its subfield. In our experiments we found a number of 5-10 top-level subfields depending on \qna data set combined with a tree depth of 3 most practical.

As the proposed knowledge stratification is a time intensive task, that relies on expert knowledge, we utilized ChatGPT as a state-of-the-art LLM to generate the knowledge hierarchy and to sort each question into the best fitting subfield. To do so, we facilitate the ChatGPT API to initially obtain the overarching subfields by prompting it to \emph{'provide a list of the 5-10 main topics of'} the knowledge field captured by the given \qna data set. Each obtained topic then represents a first-level subfield. To further stratify these subfields, we have been inspired by the work of Welbl et al. who have chosen study textbooks as base for the SciQ natural sciences dataset, since such textbooks are relevant and linguistically tailored towards a student audience~\cite{welbl-etal-2017-crowdsourcing}. In our scenario, we focus on the outline of such textbooks, which meaningfully structure the covered subject area. However, in contrast to Welbl et al., who have reviewed particular books, we again make use of ChatGPT by prompting it to \emph{'provide an outline for a textbook'} about the given subfield. Since the generated textbook outline contained an introductory and a concluding chapter in all reviewed cases, we decided to automatically drop the first and the last chapter from the obtained list if it contained a respective keyword like 'Introduction' or 'Conclusion'. Thus, with carefully crafted prompts (see Supplement for details), we are able to obtain subfield  hierarchies, from which we can build up or knowledge hierarchy data structure. All these steps are fully automatized, and the respective implementation, which is detailed in Section~\ref{sec:implementation}, will be open sourced upon publication of this paper. Nevertheless, in some cases users might want to change the obtained hierarchy. Accordingly, we allow for changes during any step during the hierarchy generation, or directly of the final hierarchy, before the questions are associated with subfields.

Since the suggested LLM-based knowledge hierarchy generation approach depends on SOTA LLMs, it is obvious, that these should be checked by expert users before being shipped with \qna datasets. Nevertheless, when considering the time it takes to curate a new \qna dataset, this verification is negligible. In the hierarchies created for this paper, experts could not spot any errors.

\section{Possible Extensions}
Besides the main visualization layout, and the two essential quantities, i.e., accuracy and number of questions, we believe that \ours are also useful to visualize further information relevant for the evaluation of LLMs. In this section, we would like to entertain this idea and outline three such examples, in order to inspire the development of future extensions to be integrated into \ours.

\noindent\textbf{Knowledge classification.} According to Tamkin et al., one central question regarding LLMs, is whether they exhibit \emph{understanding}~\cite{tamkin2021understanding}. Therefore, to assess the quality of LLMs or to find out the errors in responses or hallucinations within the responses, we should consider the dimensions of learning, which range from conceptual understanding, over creativity to problem solving~\cite{suskie2018assessing}. As, these dimensions are also highly relevant for human learners, in the past, several taxonomies for classifying learning content, have been proposed. Bloom's Taxonomy is perhaps the most widely known and used classification system for knowledge questions~\cite{anderson2001taxonomy}. It is a hierarchical framework that classifies educational learning objectives into six different levels of complexity and specificity: Remembering, Understanding, Applying, Analyzing, Evaluating, and Creating. It has been used extensively in education, from K-12 classrooms to higher education and professional development. While several other such taxonomies exist  (e.g., Structure of Observed Learning Outcomes (SOLO) Taxonomy, Webb's Depth of Knowledge, Fink's Taxonomy of Significant Learning, and Marzano's Taxonomy), due to its widespread use, we have decided to choose Bloom's Taxonomy to classify the type of knowledge gain for each question, contained in the processed \qna datasets. This can be done by analyzing the question and identifying the specific cognitive processes involved, based on which the question can be categorized. For example, a question such as 'What is the definition of photosynthesis?' would be classified as a remembering question. A question such as 'What are the causes of climate change?' would be classified as an analyzing question. A question such as 'How would you design an experiment to test the effects of different fertilizers on plant growth?' would be classified as a creating question. Thus, we consider the dimensions of learning by classifying all questions according to Bloom, and stratifying our evaluation also based on Bloom's levels. To visualize Bloom's taxonomy in the context of the evaluated LLM , we display the accuracy for each level in a pyramidal fashion, as it is usually chosen when displaying Bloom's taxonomy (see Figure~\ref{fig:teaser} (bottom)). Naturally to judge the reliability of this classification it is again important to estimate how many questions are associated with each Bloom level. To do so, we employ the same blue noise visualization as we have used to display the number of questions per subfield, while each point accounts for 1\% of all samples to facilitate the readability.

\noindent\textbf{Hallucination depiction.}  As we believe that \ours can be a great tool to spot misinformation, we introduce a simple hallucination score which we use for visualization demonstration. In this context, we consider the models self-assigned difficulty rating of each processed question on a 5 point Likert scale and compare it . We do this based on the assumption that wrong answers on self-assessed easy questions, are a strong indicator for hallucinations, as we expect the model to express uncertainty in cases, where it realizes that it faces a difficult question. This is also in line with Ji et al.'s definition of an AI hallucination as a confident response by an AI that cannot be grounded in any of its training data~\cite{ji23hallucinations}. To then obtain a measure of hallucinations, we consider the ratio function $f_r$ of average accuracy over self-assessed difficulty level. While it is difficult to make general assumptions about $f_r$, its function values can be expected to decrease monotonically with increasing difficulty. Since $f_r$ is discrete, we use the monotonicity index for discrete functions~\cite{coble2009identifying} to compute the monotonicity of $f_r$:

\begin{equation}
\label{eq:hallucination}
    s_h = \frac{1}{N-1}\sum_{i=0}^{N-1}sign(x_{i+1}-x_i),
\end{equation}

\noindent where in our case $N=5$ is the number of observations and $x_i$ the average accuracy for difficulty level $i$. From this follows $s_h\in[-1,1]$ with $s_h=-1$ being a strictly monotonically decreasing function. So, for our visualization purposes, we normalize $s_h$ to lie in the range $[0,1]$, where $0$ stands for no hallucination, while $1$ stands for maximal hallucination.

We want to emphasize, that the main contribution of our paper is the visualization part of \ours, and that the derived hallucination score is merely a tool to showcase the capabilities of our visualization, rather than a production ready LLM evaluation metric. We would leave the development of these open for NLP researchers, and would be happy to see them replacing this score in \ours with other metrics derived in literature~\cite{ji23hallucinations}.

As hallucinations are considered a major risk for LLMs~\cite{weidinger2021ethical}, we have also reserved some screen real estate to visualize additional information hinting towards these. 
To do so we use a smiley icon with concentric circles near the head of each subfield, which fill state represents the degree of hallucination for the given subfield.

\noindent\textbf{Difficulty rating.} Finally, in Figure~\ref{fig:teaser}, we also show how the difficulty rating and the models response time can be visualized in this manner. To depict response time we suggest to use a speed gauge, which shows a value normalized over all displayed questions. To depict question difficulty, ranked on a 5 point Likert scale, we have instead opted for a discrete progress bar to facilitate an easy distinguishability of each icon.

\section{Implementation Details}\label{sec:implementation}
Within this section, we briefly describe how we have implemented \ours. Due to the widespread use of Python in the ML community, we have also used it to implement \ours. We will first describe, how \qna datasets can be imported and stored, before we describe how we obtain results for these from the LLMs. Finally, we will provide some details on implementing the visualization. All generated data and source code, for processing data and generating \ours, is available on GitHub: \url{https://github.com/*****}\footnote{Data and source code will be provided in case of acceptance.}

\subsection{\qna Processing}
To leverage automatic processing of different \qna datasets in the visualization pipeline, they first need to be transformed into a common data structure. To define what information fields are necessary we combined the knowledge obtained from probing a large number of \qna datasets, whereby we for instance considered the HELM dataset~\cite{liang2022holistic} and the datasets provided by the University of Freiburg's Algorithms and Data Structures Group\footnote{\url{https://github.com/ad-freiburg/large-qa-datasets}}. Based on these reviewed datasets, our data structure contains fields for the question text, short and long versions of the answer, context information, the question type, fields for Bloom classification and difficulty rating, the topic of the question, and if applicable the answer options. Depending on the raw unprocessed dataset, not every field must necessarily be filled. As such, the type of content for short and long answers as well as for the topic and context differs depending on question type and available information. The short answer may be an index in case of multiple choice questions, a single word where questions are expected to be answered in such a way. The long answer consequently can be the text of the correct choice in multiple choice datasets, a short answer with reasoning, or simply a long answer for questions expecting such an answer. The topic field is solely dependent on the available information of the dataset. If no more information is provided it is the general topic of the original dataset. Else it can contain a general topic of each question independently or a hierarchical list of topics and subtopics for each sample. The field for context information can either contain information which clarifies the setting of the question and may be required for a correct answer, or a reference text which contains the correct answer~\cite{welbl-etal-2017-crowdsourcing}. The fields for the Bloom classification and the difficulty rating are usually not available for a given dataset and contain the LLM-assessed values as discussed in Section~\ref{sec:knowledgestratification}. Question import is in most cases done by mapping variously named data fields of the original datasets to the corresponding fields in our internal data structure. In rare cases the original data needs to be processed with pattern matching techniques to split the data within a single field in the original data structure to obtain data for multiple fields in the target structure. With the described procedure we have processed the PubMedQA, SciQ, MedMCQA, US Bar-MBE, and the four knowledge-based \qna data sets from the HELM library (MMLU~\cite{hendrycks2020measuring}, NaturalQA~\cite{kwiatkowski-etal-2019-natural}, OpenbookQA~\cite{mihaylov-etal-2018-suit}, WikiFact~\cite{petroni-etal-2019-language}), which we will release together with the automated processing scripts upon publication of this paper.

\subsection{LLM Interface}
To obtain correct and independent results in a visualizable and informative structure, the imported data needs to be processed in three steps. First we use the method proposed in Section~\ref{sec:knowledgestratification} to stratify the knowledge field, resulting in a hierarchical, tree-like structure which forms the visualization basis of \ours. Next we incorporate more information about each question by prompting the LLM to be evaluated to provide a self-assessed difficulty rating, and we further obtain a Bloom level classification~\cite{anderson2001taxonomy} as described above. Finally, we need to obtain the actual information to be visualized with \ours and thus need to obtain the answers for each question for each evaluated LLM. All of this happens also in an automated manner.

The prompting for question answering follows some simple design rules. First of all, to keep the results as comparable as possible with respect to usability of the models, we kept the respective prompts for each model as similar as possible and did not engage in any finetuning of single models. The general setup of these prompt follows a few-shot approach, where each prompt consists of three samples with a known and given answer and one sample without an answer, which has to be generated by the model. The formulation is such, that the generated answer would naturally complete the prompt. For details and examples on the prompt setup, refer to the Supplement. The only exception to this rule is given by the size restrictions on the input prompt for different models. As such ChatGPT and GTP-3~\cite{brown2020language} can process up to 4,096 tokens as input and response combined. BLOOM~\cite{scao2022bloom} can take only up to 1,000 input tokens when using the Huggingface API. The offline used models GPT-2 and LLaMA-13B~\cite{touvron2023llama}, are restricted to a context of 1,024 and 2,048 tokens respectively. Depending on the size and availability of context information and the possible size of the prompt, we then first clip the context information of the few-shot examples in cases where it is not necessary to answer the question and if the restrictions are still not met, we lower the number of few-shot examples. For each dataset the examples used for few-shot prompting are constant and removed from the evaluated set of answers. For all of these tasks, we ensure a session reset of the respective LLM model, to remove possible bias stemming from past requests.

\subsection{Visualization}
As \our's main application is communicating LLM performance, our implementation goal was to be able to use \ours in a wide range of contexts, reaching from scientific publications to webpages. Accordingly, we have opted for using SVG as a salable and universal realization for \ours, such that they can be used in many contexts. As we want ML experts to easily generate and configure \ours, we have realized a Python implementation that can generate \ours. While this makes the integration of interaction paradigms a bit more cumbersome, as compared to for instance a D3 implementation, we felt that the ease of use and adaptations outplay this downside.

\section{Use Cases}\label{sec:usecases}

\begin{figure*}[t]
\centering
    \includegraphics[trim={15 90 15 90},clip,width=\textwidth]{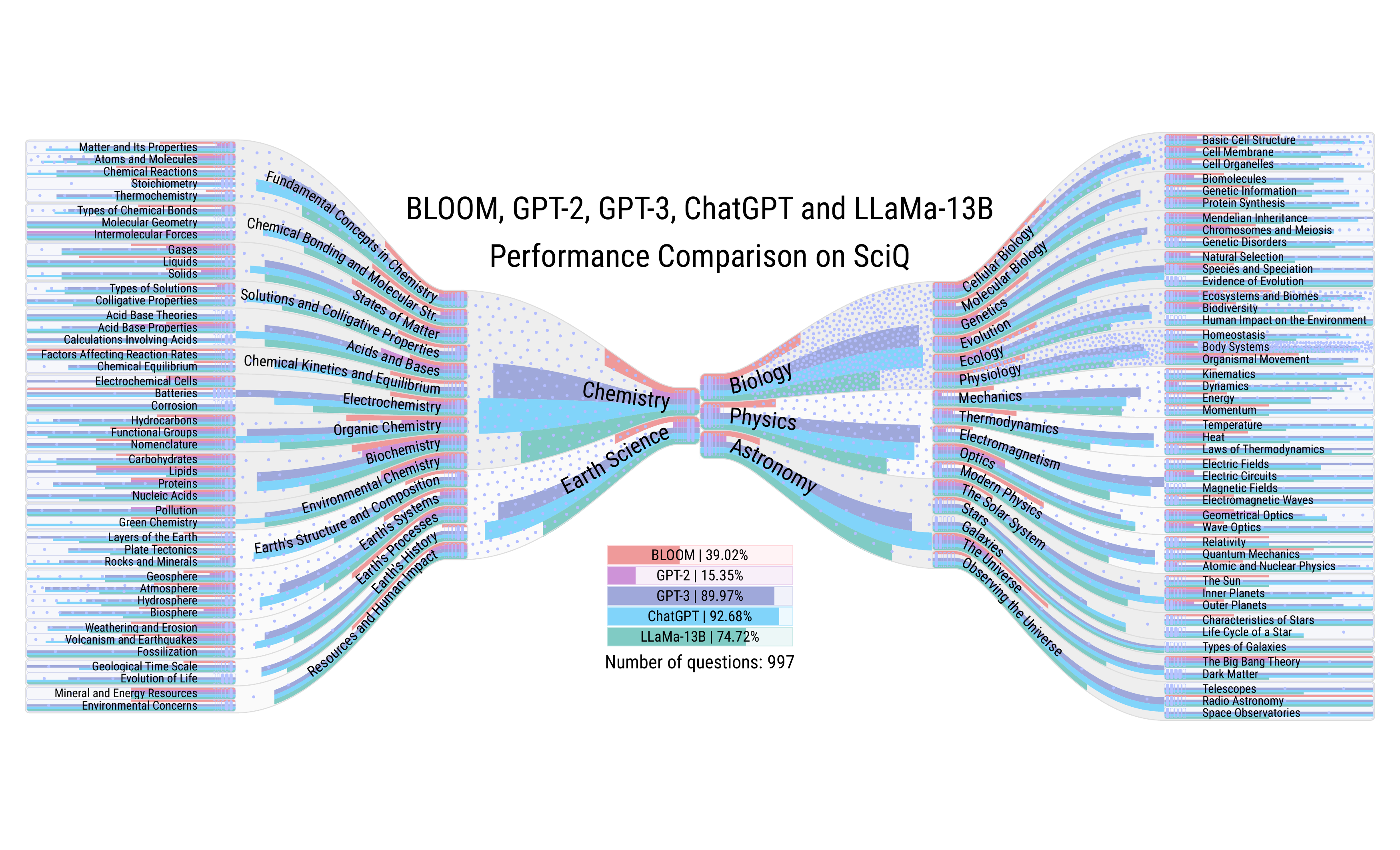} 
    \caption{Comparison of BLOOM, GPT-2, GPT-3,and LLaMa-13B on the stratified SciQ natural sciences \qna test set. Bars show model accuracy, blue noise number of questions, and discrete progress bar icons model-agnostic difficulty rating - each aggregated per knowledge hierarchy level.}
    \label{fig:multimodel}
\end{figure*}

\begin{figure*}[t]
\centering
    \includegraphics[trim={15 90 15 120},clip,width=\textwidth]{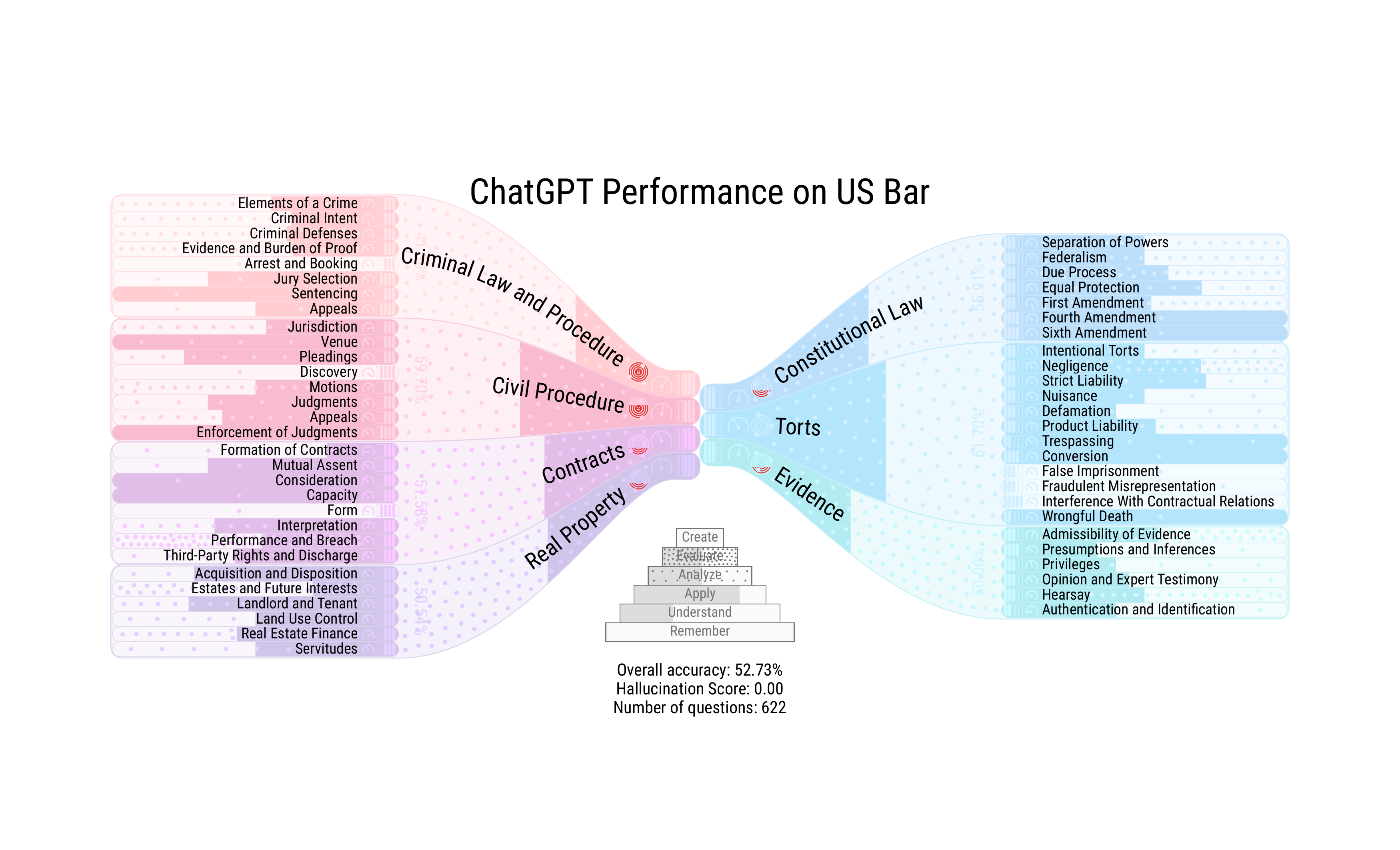} 
    \caption{\our for ChatGPT's stratified results on the US Multistate Bar Examination questions.}
    \label{fig:usbar}
\end{figure*}

\begin{figure*}[t]
\centering
    \includegraphics[trim={0 90 0 110},clip,width=0.8\textwidth]{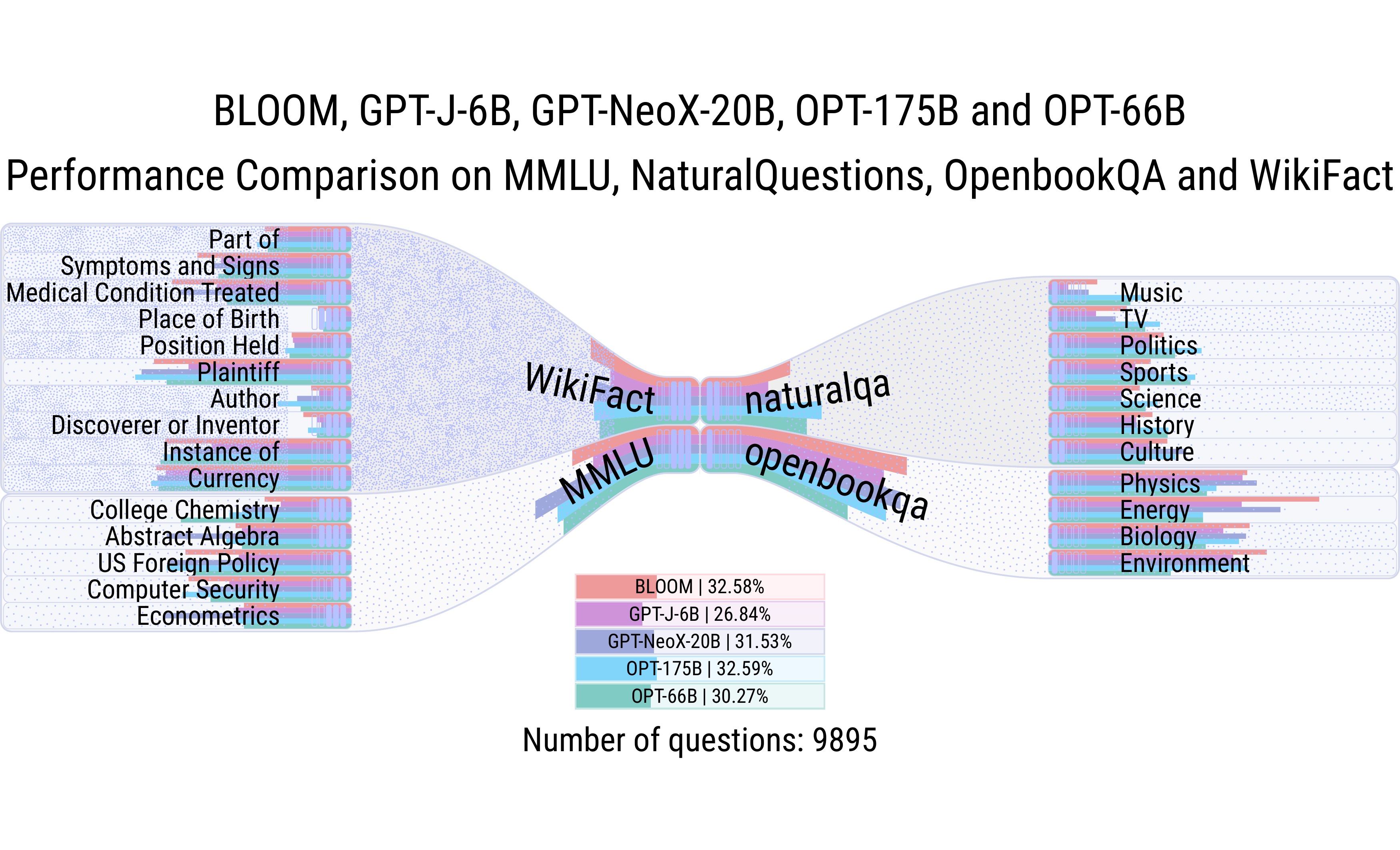} 
    \caption{LLMMap of several datasets of the HELM library for BLOOM, GPT-J-6B~\cite{gpt-j}, GPT-NeoX-20B~\cite{black2022gpt}, OPT-175B and OPT-66B~\cite{zhang2022opt}.}
    \label{fig:helm_data}
\end{figure*}

In this section we demonstrate the applicability of \ours by discussing different use cases. 

\noindent\textbf{Single-model use cases.} To get a better insight on the performance of a single model, we show the single-model use case of \ours in Figure~\ref{fig:teaser}. One thing to notice from the displayed knowledge stratification is, that ChatGPT performs significantly different for different knowledge fields in the same dataset. As such its accuracies on the first level of our knowledge hierarchy range from 59\% to 78\%, hinting that especially for genetics, regenerative medicine and infectious diseases further improvements would be required. With the visualized difficulty level assessed by ChatGPT for these questions, we can also observe that the model's perception of the questions difficulty does not always line up with its performance. For example the only subfield in which all questions are perceived as most simple is \emph{Medical Ethics and Responsible Conduct of Research} under \emph{Medical Imaging Research}. The accuracy of ChatGPT however is 0\% in this field. In stark contrast to that, the complete field of \emph{Innate Immunity}, as well as all its children, are perceived as (very) difficult by the model. Its score however shows a perfect 100\%. This is further backed by ChatGPT's hallucination score on PubMedQA, which we compute according to Equation~\ref{eq:hallucination}. With $s_h=0.25$ the hallucination score is suggesting a strong misconception by ChatGPT about its own abilities. Furthermore, while ChatGPT seems to be proficient in accessing its own capabilities in cancer research, it completely fails to do so in immunology research. In Figure~\ref{fig:usbar} we show ChatGPT's stratified results on the US Multistate Bar Examination questions. While it first appears that ChatGPT shows on average no hallucinations, the stratified visualization allows for a deeper insight into this. Here we can see, that the model actually exhibits strong hallucinattion factors for the fields of criminal law and procedure as well as civil procedure, whereas it seems completely free of hallucinations when regarding torts.

\noindent\textbf{Multi-model use cases.} In Figure~\ref{fig:multimodel} we show the performance comparison of multiple models on the test split of the SciQ dataset with \ours. The first observation we can make is about the general performance, where ChatGPT and GPT-3 perform best, but also LLaMa-13B performs surprisingly well given its relatively small size (13B vs. 175B parameters in ChatGPT). \ours stratified visualization further exposes that ChatGPT, GPT-3 and on the deepest knowledge hierarchy level, also LLaMa-13B, even perform on some subfields with a perfect accuracy. However, the low density of the blue noise dots in some of these subfields, signals caution as these subfields contain very few questions, e.g., Astronomy. So the more relevant information we can see in our \our is that ChatGPT, while scoring the best overall accuracy, is not for each subfield the best scoring model. Most notably, and with a rather high significance, this is true for Physics, where GPT-3 is not only better on average across the entire field, but even for the next subfield level is on par or better in all but one case than its successor model ChatGPT. In Figure~\ref{fig:helm_data} we apply \ours to compare several state-of-the-art LLMs on four knowledge-based \qna data sets from the HELM benchmark. Here we note that also TruthfulQA~\cite{lin-etal-2022-truthfulqa} as well as an ablation on HellaSwag~\cite{zellers2019hellaswag} are listed as knowledge tasks on the HELM website, however the HellaSwag ablation is just a transformation of a commonsense inference task into a multiple choice framework and TruthfulQA is targeted towards detection and leveraging of misconceptions, while we are interested in the evaluation of factual knowledge. Thus, Figure~\ref{fig:helm_data} shows a multi-dataset \our for the MMLU, WikiFacts, NaturalQA and OpenbookQA datasets. Since response data for this analysis has been scrapped from HELM, we report the results for some of the there available models, i.e., BLOOM, GPT-J-6B~\cite{gpt-j}, GPT-NeoX-20B~\cite{black2022gpt}, OPT-175B and OPT-66B~\cite{zhang2022opt}. The stratified visualization of WikiFact allows for the insight on the generally lower knowledge of all models with respect to questions regarding specific humans, as the accuracy on the fields \emph{Position Held}, \emph{Place of Birth}, \emph{Author} and \emph{Discoverer or Inventor} is significantly lower than on all other fields. We also observe a significantly different relative model performance between tasks. GPT-NeoX performs significantly better on MMLU than all other models and also second best on OpenbookQA, despite being moderately sized. At the same time, it only scores the second and third place on NaturalQA and WikiFact respectively. This is likely caused by MMLU and OpenbookQA being multiple choice questions, while the other two are free-form questions.

\section{Qualitative Evaluation}\label{sec:evaluation}
To obtain feedback on \ours from LLM users and experts, we have conducted two evaluation studies. First, an informal interview session with three medical researchers, in order to obtain feedback on the usage scenarios in application domains possibly using LLMs (see Subsection~\ref{subsec:medicaleval}). Second, a structured interview session with four language model developers, where we asked them specific questions in order to assess how useful \ours can be for LLM development and training (see Subsection\ref{subsec:mleval}).

\subsection{LLM User Evaluation}\label{subsec:medicaleval}
Within an informal interview session, we have confronted three medical researchers from the areas of radiology, nuclear medicine and medical radiation physics (between 11 and 27 years of medical research experience) with our analysis of ChatGPT’s performance on PubMedQA (see Figure~\ref{fig:teaser}). While all of them knew about ChatGPT and to some extend about LLM in general, two already had made first medical experiments with ChatGPT and where in general impressed by its text processing capabilities, while being at the same time cautions regarding its factual knowledge. As none of the three researchers knew about the PubMedQA dataset, we provided a brief introduction, detailing the question types and available annotations. All three appreciated the availability of the dataset, and where curious to learn about ChatGPT’s performance on this dataset.

Confronting the three researchers with \ours for ChatGPT on PubMedQA sparked engaging and explorative behavior. All three were interested to know, which medical fields were covered, and looked in particular into subfields representing their own area of research. After an initial exploration, we have explained the process, which we have applied to obtain the knowledge hierarchy, and asked them if they can spot subfields which are unexpected or missing. After giving it some thought, they were realizing, that some relevant basic medical subfields, such as orthopedics, dermatology, ophthalmology or internal medicine in general, were missing or underrepresented. While they could not spot any subfield, they would have considered unexpected, they pinpointed one example, where they would have expected the hierarchy to be different, i.e., ‘Computed tomography (CT) imaging’ being a subfield of ‘Radiographic Imaging’. After a small discussion they though acknowledged that this could also be a valid interpretation. Nevertheless, asked which alternative approaches they could imagine to obtain a knowledge hierarchy for the field of medicine, they suggested to scrape information from university hospital websites and use the underlying department structure. To follow up on this idea, we have prompted ChatGPT with ‘Which medical departments should a medical university hospital contain?’, and obtained the following list, which in fact contains the missing subdisciplines: Emergency Medicine, Internal Medicine, Surgery, Pediatrics, Obstetrics and Gynecology, Radiology and Imaging, Pathology, Neurology, Cardiology, Orthopedics, Dermatology, Ophthalmology, Oncology, Psychiatry, Rehabilitation Medicine, Infectious Diseases, Anesthesiology. We believe that this nicely shows, that modern LLMs have the capability to provide knowledge hierarchies while the output though highly depends on the prompting. For the future, this would mean that knowledge hierarchy prompting should further be investigated, and that the prompts should probably also be communicated with the knowledge hierarchy.

Regarding the visualization part of \ours, after a brief (<1 minute) introduction, all three researchers could directly understand how to interpret the visualization, and how to derive and compare the accuracy numbers for the different subfields. They were able to compare different subfields and in particular curious to know how ChatGPT performs on their own area of research. Since initially we discussed how the amount of questions per subfield can be seen, they could successfully interpret our \qna sample encoding as they immediately realized and mentioned that overall ethical questions seem to be less frequent in the data. So overall, we believe that this initial and informal interview shows, that application experts see the need for better understanding the knowledge capabilities of LLMs. Furthermore, we conclude that \ours do not only provide an engaging mechanism to visually explore these capabilities, but that the visual encoding is also intuitive and understandable. Finally, the separation of knowledge hierarchy generation and visualization provides needed flexibility, which allows for the adaptation of knowledge hierarchy generation for dedicated use cases. One of the medical researchers for instance stated, that he would be interested in seeing a knowledge hierarchy which is focused on his area of research and possibly also containing medical questions beyond PubMedQA.

\subsection{LLM Developer Evaluation}\label{subsec:mleval}
To further investigate the benefits of \ours in the context of LLM development, we conducted a more technical evaluation with four academic researchers in the form of structured one-on-one interviews. The four participants were selected due to their familiarity with LLMs (between 1 and 2 years experience working with LLMs). Each participant was shown three figures showing applications of \ours on the SciQ dataset. Two showing the performance of ChatGPT with and without empty leaf nodes, and one multi-model case comparing the performance of GPT-3, ChatGPT, and BLOOM on the same dataset. The interviews comprised of open ended questions, and the participants have not been compensated for the interviews.

We first asked the participants whether they could draw more information from \ours as compared to the conventional display of model accuracies and if this information is helpful. To this we received in general positive responses from every participant with some additional feedback regarding the visualization of the multi-model case, as well as the number of samples in the leaf nodes, which we further incorporated into our visualization designs as shown in this paper. The participants mentioned the combined display of information in a single figure, the stratification of the data domain as well as the containment of additional quantifiers like the difficulty, response time and Bloom taxonomy as valuable. Next, we asked the participants about the comprehensibility of \ours. All participants found the visualization easy to comprehend, apart from minor flaws perceived in the visual design which we were able to incorporate into our final design. While all found it especially helpful, that the multi-model \ours quickly convey the relative model performances for individual subfields, one participant noted that he would like to see \ours in addition to the conventional model accuracy bar charts, as \ours are harder to digest than those due to the additional information. We took also this suggestion into account, and added the overall performance in the model legend as seen in Figure~\ref{fig:multimodel}. We then asked the participants to imagine an LLM application, and if \ours would help them in deciding when to trust and when not to trust the visualized models. All participants noted that the stratified visualization helps in understanding the model performance in the given subfields. Three of the four participants also noted that on one hand the visualization of the number of questions in a given node helps gaining trust to the model, while also raising a feeling of caution towards such areas where only a few samples are contained. The next two questions revolved around the data domain. First we asked if \ours help in understanding the \qna's knowledge field. To this we received only affirmative answers hinting on the visualized subfield information. On the question if this also helps identifying content driven, i.e. knowledge based bias we received mixed responses. While two answers were strictly affirmative due to the stratified visualization of accuracies and the information of sparsely or unpopulated subfields, one participant noted that it would be helpful if a higher number of samples would be available, and another one mentioned that this question is difficult to answer, as the training data is not known. Lastly we asked if they prefer a visualization with or without the empty leaf nodes. While all participants preferred the display of empty leaf nodes due to the insights on missing data to evaluate these areas, they also strictly found the visualization without empty leaf nodes to be easier to comprehend and would thus prefer it in cases where \ours become larger.

Finally, we received some general feedback, where several participants pointed out that they found the visualization unique and novel, and that they would like to use \ours in their own research publications.

\section{Conclusions and Future Work}\label{sec:conclusions}
In this paper, we have introduced \ours as a visual metaphor for the stratified evaluation of one or several LLMs on one or several \qna datasets. \ours provides a stratified evaluation of LLM performance on knowledge-based \qna datasets, and the results obtained in our qualitative user studies indicate, that \ours enable LLM users to better assess and compare the capabilities of existing LLMs, while at the same time supporting LLM developers with model improvements. These aspects are especially important in critical domains, where accuracy and reliability of LLMs are most crucial. We believe, that our work is well inline with current research initiatives, i.e., the HELM initiative~\cite{liang2022holistic} and OpenAI's Evals benchmark repository. We further see \ours as an ideal extension to these frameworks, as it allows for a more stratified evaluation of the provided benchmarks.

While we have developed \ours for the evaluation of LLMs, which is therefore also the major focus of this paper, \ours could also be used to provide an overview over \qna datasets only. In these scenario, one would not display any LLM performance information, but rather use \ours to display the question distribution of a \qna dataset over subfields, in order to inspect and decide which subfields are under- or over-represented. In this scenario, we would also visualize those leaf nodes of our knowledge hierarchy, which do not contain any questions, as these potentially depict topic areas which are considered important, but have no questions associated with them.

For future work we see several possible endeavors. Most obvious would be the development of \ours into an interactive application. While we have deliberately decided not to do so at this point, since we focus on the communication of evaluation results for scientific papers and elsewhere, we could imagine other scenarios which benefit from such interactions supporting an interactive exploration. Additionally, we believe that the area of knowledge stratification should get further attention. While the proposed approach works well for educational datasets, applying it to more trivia-like questions, incorporating subfields ranging from music over gossip to politics does not benefit from our textbook approach. Furthermore, in this paper we solely focus on the stratified evaluation of knowledge capabilities. Another large suite of benchmarks is available to investigate language understanding of LLMs. In the future, we believe it would be interesting to derive appropriate hierarchies for this context, e.g., following the radial example by Wang et al.~\cite{wang2022self}, and see how \ours can contribute in this area.


\bibliographystyle{abbrv-doi-hyperref}
\bibliography{llmmaps}

\end{document}